\documentclass[letterpaper, 10 pt, conference]{ieeeconf}  

\IEEEoverridecommandlockouts                              

\makeatletter
\let\NAT@parse\undefined
\makeatother

\usepackage{cite}
\usepackage[caption=false,font=footnotesize]{subfig}
\usepackage[hidelinks]{hyperref}
\usepackage{amsmath,amssymb,amsfonts}
\usepackage{algorithmic}
\usepackage{graphicx}
\usepackage{textcomp}
\usepackage{xcolor}
\usepackage{subfig}
\usepackage{tabularx}
\usepackage{booktabs}
\usepackage{multirow}
\usepackage{siunitx}
\usepackage{eso-pic}
\AddToShipoutPictureBG*{%
  \AtPageLowerLeft{%
    \put(\LenToUnit{0.5\paperwidth},\LenToUnit{6mm}){%
      \makebox[0pt]{\parbox{0.86\paperwidth}{\centering\scriptsize
      \copyright~2026 IEEE. Personal use of this material is permitted. Permission from IEEE must be obtained for all other uses, in any current or future media, including reprinting/republishing this material for advertising or promotional purposes, creating new collective works, for resale or redistribution to servers or lists, or reuse of any copyrighted component of this work in other works. Presented at the 2026 IEEE International Conference on Advanced Robotics and Mechatronics (ICARM).}}}}}

\title{\LARGE \bf
Centralized Multi-UAV Exploration and 3D Reconstruction Using Single-UAV Planners
}

\author{João Félix Mendes,
Meysam Basiri, and
Rodrigo Ventura
\thanks{This work was supported by Aero.Next project (PRR - C645727867-00000066) and LARSyS FCT funding (DOIs: 10.54499/LA/P/0083/2020, 10.54499/UIDP/50009/2020 and 10.54499/UIDB/50009/2020).}
\thanks{The authors are with the Institute for Systems and Robotics, Department of Electrical and Computer Engineering,
Instituto Superior Técnico, Lisboa, Portugal. {Corresponding Author: João Félix Mendes (e-mail: joao.felix.mendes@tecnico.ulisboa.pt).}
}
}

\begin{document}

\maketitle
\thispagestyle{empty}
\pagestyle{empty}

\begin{abstract}

Extending single Unmanned Aerial Vehicles (UAVs) exploration methods to multi-UAV teams can improve coverage speed and robustness, but introduces challenges such as consistent mapping, safe navigation, and deployment strategy. In this work, we present a centralized multi-UAV exploration framework that enables the use of existing single-UAV sampling-based planners in a multi-UAV setting.

The proposed architecture allows multiple UAVs to collaboratively explore unknown environments using a shared global Truncated Signed Distance Field (TSDF) map and centralized planning. Building on the voxblox library, we adapt its mapping pipeline to support real-time fusion of depth measurements from multiple UAVs into a common TSDF representation. In addition, inter-UAV collision avoidance and robot self-filtering mechanisms are integrated into the system to ensure safe navigation and prevent reconstruction of other UAVs as static obstacles.

The framework is evaluated in simulation using four sampling-based exploration planners - RH-NBVP, KRH-NBVP, AEP, and KAEP - whose core sampling logic is preserved, with only system-level adaptations for multi-UAV operation. Experiments are conducted across multiple environments and under two deployment configurations: Joint Start (JS), where UAVs are initialized in close proximity, and Separated Start (SS), where UAVs are initialized in distinct locations. Results show that SS deployments consistently achieve faster exploration and improved coverage across all planners, highlighting the importance of the deployment strategy in multi-UAV exploration performance.

\end{abstract}

\section{INTRODUCTION} \label{sec:introduction}

Equipping Unmanned Aerial Vehicles (UAVs) with autonomous exploration capabilities enables applications like search and rescue~\cite{Dang_2020}, inspection of inaccessible environments~\cite{Petracek_2021}, and urban planning~\cite{Mendes_2026}. While advances in sampling-based planning \cite{Mendes_2026, Bircher_2016, Selin_2019, Schmid_2020} have significantly improved single-UAV exploration efficiency in complex environments, extending these methods to multi-UAV systems can accelerate exploration and improve robustness to local planning failures. However, this introduces coordination, data fusion, and safety challenges unaddressed by single-robot formulations.

In this work, we adopt a centralized multi-UAV architecture, where a central unit aggregates sensor data from all robots to compute global mapping and planning decisions. While this architecture limits scalability compared to decentralized systems, it provides a key advantage in exploration: access to global information. By maintaining a unified representation of the environment and system state, the centralized approach enables global map fusion and facilitates conflict-aware coordination across agents. Unlike decentralized approaches that rely on inter-robot communication and coordination protocols, the centralized architecture allows each UAV to make planning decisions based on globally available data. This simplifies coordination and makes the centralized architecture well-suited for studying multi-UAV exploration behavior and evaluating the extension of single-UAV planners to multi-UAV systems.

\begin{figure}[t]
    \centering
    \subfloat[Joint Start (JS)]{\label{subfig:recon_js}
    \includegraphics[width=0.48\linewidth]{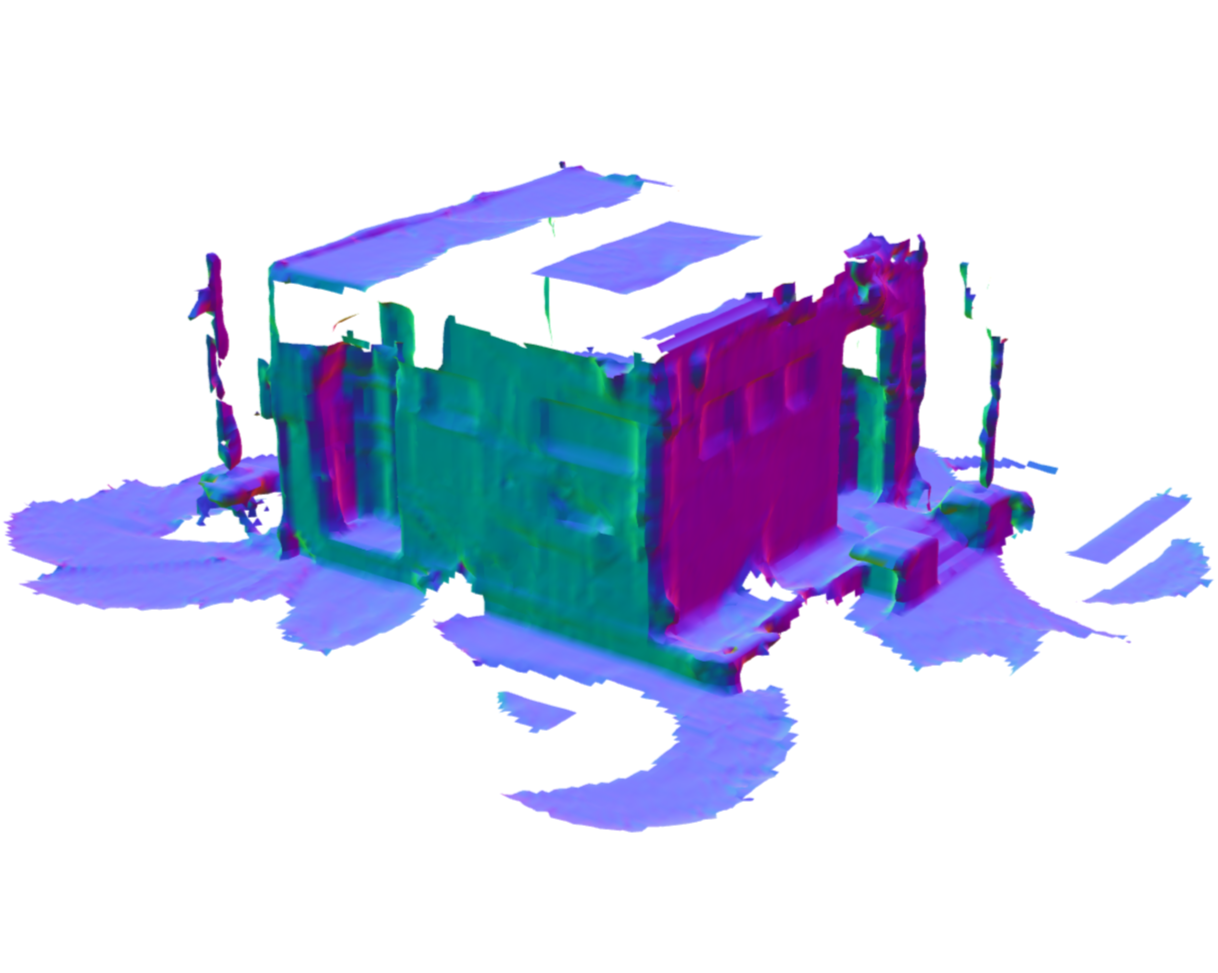}}%
    \hfil
    \subfloat[Separated Start (SS)]{\label{subfig:recon_ss}
    \includegraphics[width=0.48\linewidth]{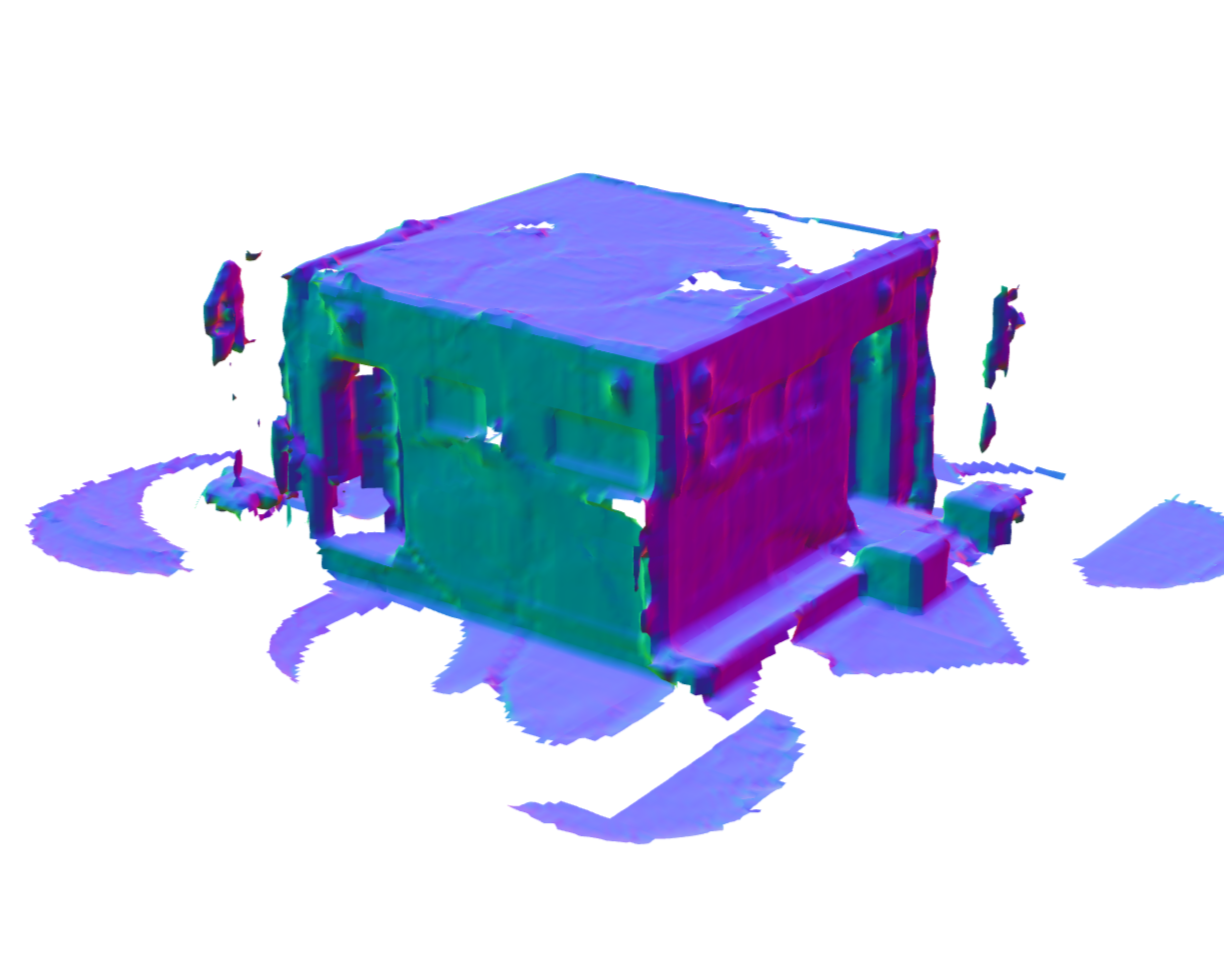}}%
    \caption{Map reconstructions generated by a three-UAV team using the KAEP planner in the Police Station environment at $t = \SI{160}{\second}$. \textbf{(a)} In JS, UAVs begin in close proximity, resulting in more overlapping coverage. \textbf{(b)} In SS, UAVs are spatially distributed, leading to broader coverage with reduced overlap.}
    \label{fig:map_reconstruction_comparison}
\end{figure}

However, bridging the gap between single and multi-robot exploration within a centralized architecture is nontrivial. Well-established exploration algorithms, such as the Receding Horizon Next-Best-View Planner (RH-NBVP) \cite{Bircher_2016}, cannot be directly deployed in multi-UAV settings without system-level adaptations, as they are designed for single-UAV operation and lack multi-robot mechanisms such as shared mapping, inter-UAV collision avoidance, and mutual UAV reconstruction prevention. In particular, extending widely adopted mapping frameworks such as voxblox \cite{oleynikova2017voxblox}, which uses a Truncated Signed Distance Field (TSDF) representation designed for single-sensor integration, requires careful handling of multi-sensor fusion to maintain map consistency. Additionally, safe operation requires avoiding inter-UAV collisions and preventing UAVs from being reconstructed as static obstacles. Otherwise, without these, UAVs may treat other robots as static obstacles, which would artificially constrain the exploration space and degrade planner performance.

In this work, we present a centralized multi-UAV exploration framework that enables the deployment of single-UAV sampling-based planners in a multi-UAV setting. Rather than proposing new multi-robot exploration algorithms or relying on explicit coordination methods (e.g., market-based task allocation \cite{Zlot_2002}), we extend existing planners through system-level integration, where fleet cooperation is achieved through implicit coordination. Specifically, UAVs operate on a shared global map and account for the planned trajectories of other agents during planning, allowing each robot to make informed decisions based on the team’s collective knowledge while reducing redundant exploration and ensuring collision-free operation. To support this architecture, we adapt the voxblox framework for real-time fusion of depth measurements from multiple UAVs into a shared map. Furthermore, we integrate inter-UAV collision avoidance and robot pointcloud filtering into the planning process, allowing multiple UAVs to safely explore without introducing dynamic obstacles into the map.

The proposed framework is evaluated using four exploration strategies - RH-NBVP \cite{Bircher_2016}, Kinodynamic Receding Horizon Next-Best-View Planner (KRH-NBVP) \cite{Mendes_2026}, Autonomous Exploration Planner (AEP) \cite{Selin_2019}, and Kinodynamic Autonomous Exploration Planner (KAEP) \cite{Mendes_2026} - applied without modification to their core sampling logic. Through extensive simulation experiments in diverse environments, we analyze the impact of different initial deployment configurations. In particular, we compare Joint Start (JS) scenarios, where UAVs are initialized in close proximity, with Separated Start (SS) scenarios, where UAVs begin exploration from spatially distinct locations. Results show that SS deployments consistently achieve faster exploration and improved coverage, highlighting the importance of deployment strategy in multi-UAV systems. The full framework is released as an open-source package\footnote{\url{https://github.com/IRSg-ARG/UAV_3d_reconstruction}}. The main contributions of this paper are:
\begin{itemize}
    \item A centralized multi-UAV exploration framework that allows collaborative exploration of unknown environments using established single-UAV planning algorithms without modification to their core logic.
    \item An adaptation of the Voxblox dense mapping framework for real-time, centralized multi-UAV operation, enabling shared TSDF mapping from multiple sensors while preserving map consistency.
    \item An experimental evaluation validating the accuracy and real-time performance of the centralized map fusion and demonstrating the impact of initial deployment strategies on exploration efficiency.
\end{itemize}

\section{RELATED WORKS} \label{sec:related_works}

The fundamental goal of the autonomous exploration problem is to find feasible paths that allow a robot to completely map an unknown scenario in real-time. Early approaches focused on frontier-based exploration, where robots iteratively move toward the boundary between free space and unknown space \cite{Yamauchi_1997}. Several extensions have been proposed to improve the method, including clustering frontier voxels to reduce computational cost \cite{Dai_2020} and adapting frontier selection for high-speed aerial exploration by constraining motion within the sensor field of view \cite{Cieslewski_2017}. While frontier-based methods are simple to implement, they are inherently limited to exploring frontier regions and may overlook informative viewpoints located deeper inside free space.

To overcome the limitations of frontier-based exploration, sampling-based approaches were proposed. These methods directly sample candidate viewpoints in free space and evaluate them using information-theoretic objectives. A prominent example is the RH-NBVP \cite{Bircher_2016}, which samples viewpoints using a Rapidly-exploring Random Tree (RRT) and executes the initial segment of the branch that maximizes an objective function. Extensions of RH-NBVP have incorporated localization uncertainty \cite{Papachristos_2017}, history-aware reseeding to escape local minima \cite{Witting_2018}, and global planning mechanisms to ensure full coverage \cite{Selin_2019}. Other sampling-based approaches include RRT*-inspired planners that maintain a single expanding tree to guarantee global exploration \cite{Schmid_2020}, as well as trajectory-centric formulations such as KRH-NBVP \cite{Mendes_2026} and KAEP \cite{Mendes_2026}, which directly integrate trajectory optimization into planning.

Extending these single-robot methods to multi-robot systems can significantly accelerate environment coverage. As a result, prior work has largely focused on designing specific coordination algorithms to distribute tasks among robots. Common strategies include market-based task allocation \cite{Zlot_2002}, utility-based frontier assignment \cite{Burgard_2005}, and decentralized collaborative planning \cite{Zhou_2023}. While effective at coordinating a team of robots, these methods typically require reformulating the exploration logic specifically for the multi-robot case. In contrast, limited attention has been given to adapting well-established single-robot exploration pipelines to multi-robot systems without modifying their core planning logic. Centralized architectures offer a pathway to extend standard single-robot planners directly to multi-robot settings by aggregating system knowledge into a central control unit. This enables the direct reuse of single-UAV sampling-based planners, such as RH-NBVP, within a multi-robot setting. 

Mapping plays a central role in both single and multi-robot exploration systems. While probabilistic occupancy grids such as OctoMap \cite{Hornung_2013} have been widely used for single and multi-robot mapping, they lack the accuracy required for precise surface reconstruction. As a result, exploration frameworks increasingly rely on dense volumetric representations based on TSDFs. Voxblox~\cite{oleynikova2017voxblox}, a CPU-based volumetric mapping library, has become standard for single-UAV exploration frameworks due to its surface reconstruction accuracy and computational efficiency. However, native voxblox is strictly designed for single-sensor operation. Extensions such as c-blox \cite{Millane_2018} introduce sub-mapping to handle large-scale drift, while voxgraph \cite{Reijgwart_2020} leverages these sub-maps to perform pose-graph optimization. While these libraries ensure global consistency in large-scale scenarios and are theoretically applicable to multi-robot scenarios through sub-map fusion, they impose significant computational overhead. This limits their suitability for real-time mapping applications. As such, lightweight mapping solutions for accurate, real-time multi-robot mapping remain limited and underexplored in the existing literature.

In contrast to prior work that proposes new exploration algorithms or explicit coordination strategies, we focus on extending established single-UAV sampling-based planners into a centralized multi-UAV framework. By adapting voxblox \cite{oleynikova2017voxblox} to support real-time multi-robot depth measurement fusion and by experimentally evaluating the impact of deployment strategies, this work provides practical insights into the behavior and performance of multi-UAV exploration systems without modifying the underlying planning algorithms.

\section{PROPOSED APPROACH} \label{sec:proposed_approach}

\subsection{System Architecture}
Scaling autonomous exploration algorithms from single-robot to centralized multi-robot systems requires addressing three fundamental challenges. First, sensor data from all robots must be fused into a shared global map. Second, motion planning must ensure collision-free operation not only with the environment but also among robots. Third, robots must not reconstruct one another, as inter-robot mapping introduces them as obstacles in the map that restrict the exploration space. 

\begin{figure}[t]
\centering
\includegraphics[width=0.49\textwidth]{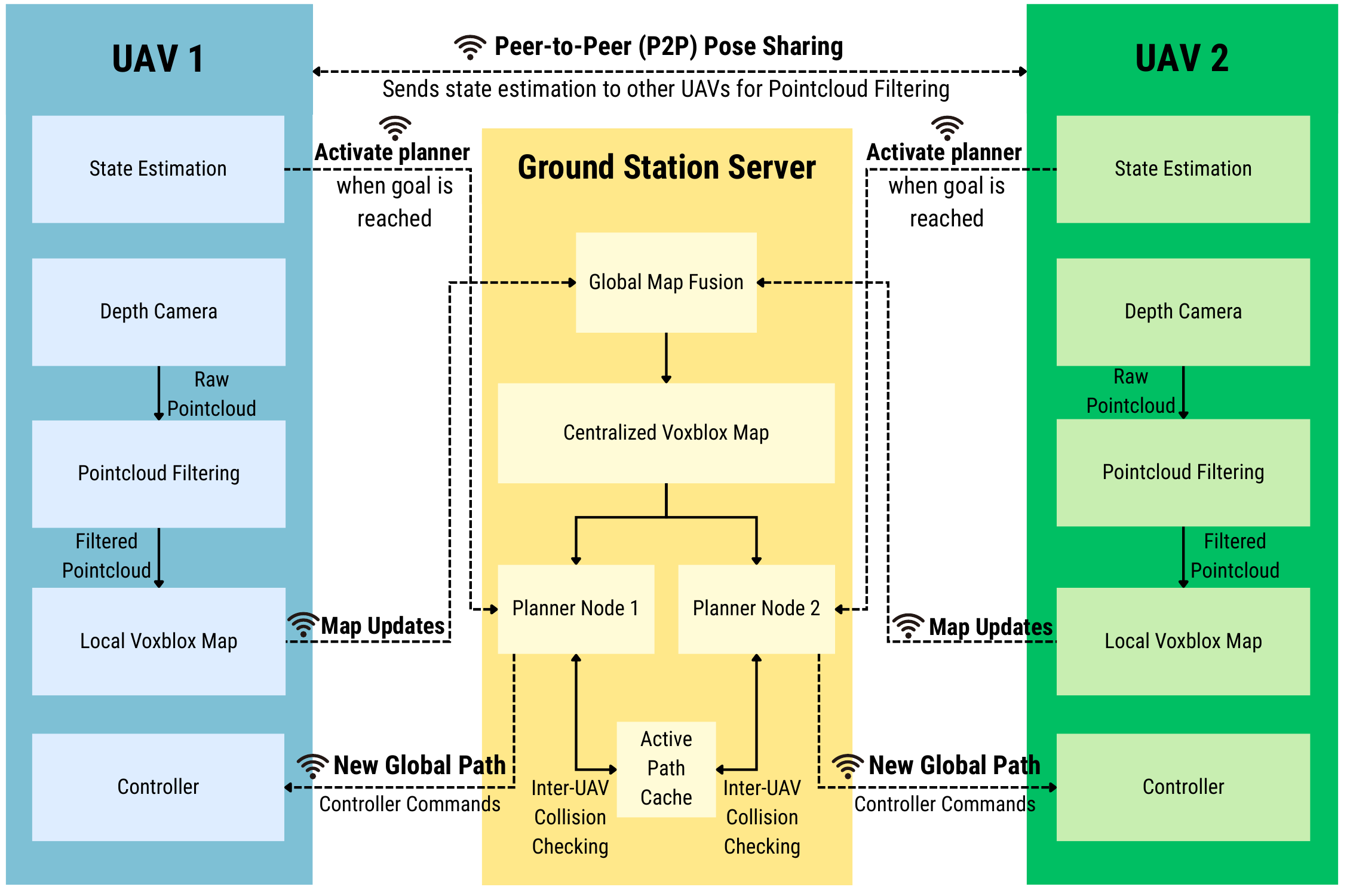}
\caption{Overview of the proposed centralized multi-UAV exploration framework for two UAVs. Color-coded blocks denote hardware boundaries (blue and green for UAVs and yellow for the central server). Each UAV performs state estimation, sensing, and local pointcloud filtering via peer-to-peer pose sharing, integrating measurements into a local sub-map. These local updates are periodically sent to the server and fused into a unified voxblox map. The planner nodes then asynchronously compute new paths upon request, utilizing the unified map and a shared active path cache for inter-UAV collision avoidance. The collision-free paths are then transmitted back to the UAVs.}
\label{fig:multi_planner_overview}
\end{figure}

Figure \ref{fig:multi_planner_overview} presents the system architecture of the proposed centralized multi-UAV framework for a team of two UAVs, explicitly dividing computation between the resource-constrained UAVs and a central ground station server. To reduce communication bandwidth and latency, high-frequency processing is performed onboard the UAVs. Each UAV receives the estimated state of other agents in the system via peer-to-peer (P2P) sharing, allowing it to filter out other UAVs from its raw depth measurements. These filtered pointclouds are then integrated into local voxblox sub-maps onboard the UAVs and periodically transmitted as incremental updates to the ground station, where they are fused into a unified voxblox map.

To enable collision-free coordination while avoiding the latency associated with continuous control streaming, motion planning is handled asynchronously. When a UAV nears the end of its current goal, it triggers a path request to the server. The server executes the underlying sampling-based planner (e.g., RH-NBVP~\cite{Bircher_2016}, AEP~\cite{Selin_2019}, KRH-NBVP, KAEP~\cite{Mendes_2026}) using the global map and a shared active path cache to account for the planned trajectories of other UAVs and avoid inter-robot collisions. The resulting collision-aware path is then transmitted back to the UAV's onboard controller for execution.

Our implementation reuses an algorithmic module from the open-source repository associated with \cite{Bircher_2016}. While the original publication is limited to single-robot exploration, the associated codebase contains experimental multi-robot structures. In particular, we adopted the core inter-robot collision avoidance logic outlined in their implementation, while substantially re-designing its integration to operate within our centralized mapping and planning framework.

\subsection{Background on Evaluated Planners}
To demonstrate the versatility of the proposed architecture, we evaluate our framework using four single-UAV exploration strategies. The core logic of these planners is briefly summarized below:
\begin{itemize}
    \item \textbf{RH-NBVP \cite{Bircher_2016}:} A receding-horizon planner that samples viewpoints using an RRT, executing the first segment of the branch that maximizes an objective function. This function balances volumetric information gain and path cost.
    \item \textbf{AEP \cite{Selin_2019}:} A hybrid planner that combines an RRT-based local planner for fast computation with a frontier-based global planner to prevent the system from getting stuck in local minima.
    \item \textbf{KRH-NBVP \cite{Mendes_2026}:} A kinodynamic extension of RH-NBVP that directly samples in the UAV's state space, producing dynamically feasible trajectories rather than geometric waypoints.
    \item \textbf{KAEP \cite{Mendes_2026}:} A kinodynamic extension of AEP, combining global frontier guidance with dynamically feasible local trajectory optimization.
\end{itemize}

\subsection{Centralized Multi-Robot Mapping}
\label{subsec:multi_centralized_map}

Voxblox \cite{oleynikova2017voxblox} is a CPU-based volumetric mapping framework that provides efficient pointcloud integration and accurate reconstruction models via the marching cubes algorithm \cite{Lorensen_1987}, making it an appropriate choice for high-quality real-time mapping and exploration of unknown environments. However, its native architecture is limited to single-sensor integration and does not directly support multi-robot mapping. 

To enable centralized multi-robot exploration, we implement a lightweight extension that allows volumetric map information from multiple robots and sensors to be fused into a shared map. Specifically, a local voxblox map is maintained for each robot in order to emulate a sub-map. The data from these local maps are periodically merged into a centralized voxblox map that serves as the global map used for planning. While physical multi-robot systems are inherently asynchronous and subject to communication delays, the proposed framework mitigates strict synchronization requirements by operating on locally integrated TSDF sub-maps. High-frequency sensor data is first integrated into local voxblox sub-maps onboard each UAV, meaning the centralized system operates on locally consistent volumetric representations rather than raw pointcloud streams. As a result, the fusion process is less sensitive to small temporal misalignments or network latency. To maintain temporal consistency across the fleet, the system assumes standard synchronization mechanisms commonly employed in multi-robot systems, such as the Network Time Protocol (NTP).

The centralized map and each local voxblox map are characterized by an underlying TSDF layer and a derived mesh layer generated via the marching cubes algorithm \cite{Lorensen_1987}. In the TSDF layer, each voxel stores the signed distance to the nearest surface and a weight reflecting the reliability of the distance measurements in the sensor frame. During map fusion, corresponding voxels from local maps are merged into the centralized TSDF by combining their weights and computing a weighted average of the stored distances. For each voxel $\mathbf{v}_i$, the combined weight and distance are computed as:
\begin{gather}
     w_{comb}(\textbf{v}_i) = w_{loc}(\textbf{v}_i) + w_{centr}(\textbf{v}_i), \\
     d_{comb}(\textbf{v}_i) = \frac{d_{loc}(\textbf{v}_i) w_{loc}(\textbf{v}_i) + d_{centr}(\textbf{v}_i) w_{centr}(\textbf{v}_i)}{w_{comb}(\textbf{v}_i)},
\end{gather}
where $d$ represents the signed distance, $w$ the voxel weight, and $loc$, $centr$, and $comb$ correspond to the local voxblox map of each robot, the current centralized map, and the combined centralized map. This formulation incrementally integrates new observations while preserving the accumulated confidence of prior measurements.

Compared to other sub-mapping approaches also based on voxblox, such as c-blox \cite{Millane_2018} and voxgraph \cite{Reijgwart_2020}, this centralized fusion strategy prioritizes computational efficiency, making it suitable for real-time mapping in multi-UAV exploration scenarios.

\subsection{Multi-Robot Collision Avoidance} 
\label{subsec:multi_collision_avoidance}

Safe navigation in a multi-robot system requires avoiding both static obstacles and other robots in the system. Static collision avoidance follows a similar strategy to established approaches \cite{Bircher_2016, Selin_2019, Mendes_2026, Schmid_2020}: the planner queries the TSDF layer of the centralized map, and compares the distance between planned path points and the nearest surface against a minimum safety distance threshold $d_{collision}$. 

To prevent inter-robot collisions, each robot shares its planned path with the others. During path generation, candidate paths are checked against the received paths of other robots by computing the minimum distance between their respective points. 

Consequently, any candidate path whose distance to either the static environment or another robot falls below $d_{collision}$ is marked as non-traversable and pruned from the planning tree, ensuring that only collision-free paths are retained for evaluation.

\subsection{Multi-Robot Pointcloud Filtering} 
\label{subsec:multi_ptcloud_filter}

In multi-robot exploration, robots may observe one another with their perception sensors. If such measurements are integrated into the map, the robots are reconstructed as occupied space. Since robots are dynamic, this results in persistent false obstacles that remain in the map after the robots move, unnecessarily constraining future path planning and reducing the traversable free space.

To address this issue, a pointcloud filtering step is applied before map integration. For each robot, a spherical region centered at the estimated poses of the other robots is defined, with radius equal to the collision distance $d_{collision}$. All pointcloud data within these regions is removed before integration into the centralized map, preventing robots from being reconstructed as static obstacles. Since $d_{collision}$ is also enforced during planning as a minimum clearance to static surfaces, robots never operate closer than this distance to real obstacles, ensuring that the filtering does not remove valid measurements in the reconstructed map.

\begin{figure}[t]
    \centering
    \subfloat[Experimental setup: Two UAVs face each other, spaced 4 meters apart.]{\label{fig:drones_setup}
        \includegraphics[width=0.22\textwidth]{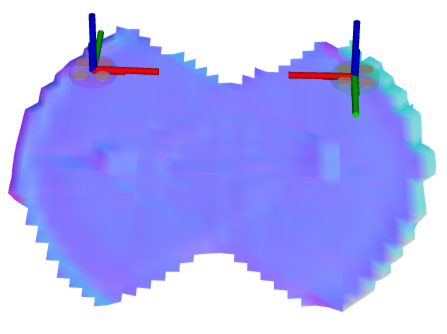}} \\
    \vspace{0.5em}
    \subfloat[TSDF visualization from the left UAV using unfiltered pointclouds.]{\label{fig:unfiltered_left}
        \includegraphics[width=0.18\textwidth]{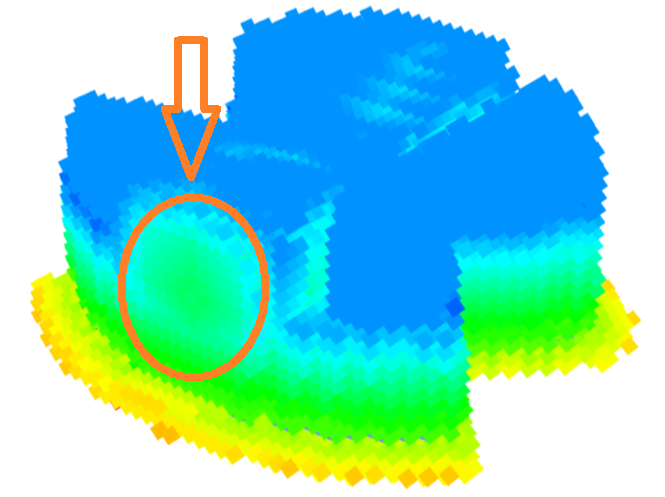}}
    \hspace{1em}
    \subfloat[TSDF visualization from the right UAV using unfiltered pointclouds.]{\label{fig:unfiltered_right}
        \includegraphics[width=0.18\textwidth]{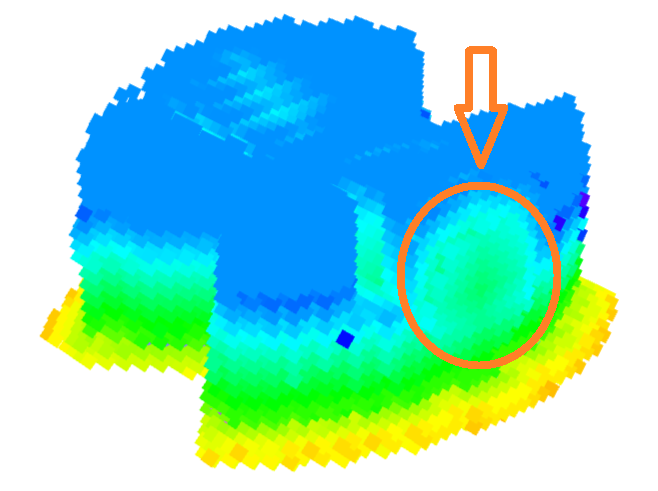}} \\ 
    \vspace{0.5em}
    \subfloat[TSDF visualization from the left UAV with pointcloud filtering enabled.]{\label{fig:filtered_left}
        \includegraphics[width=0.18\textwidth]{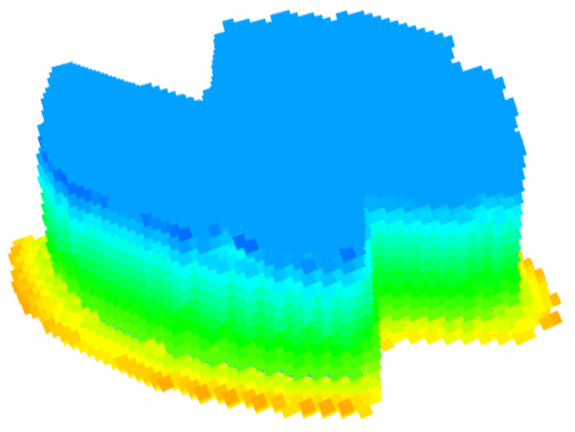}}
    \hspace{1em}
    \subfloat[TSDF visualization from the right UAV with pointcloud filtering enabled.]{\label{fig:filtered_right}
        \includegraphics[width=0.18\textwidth]{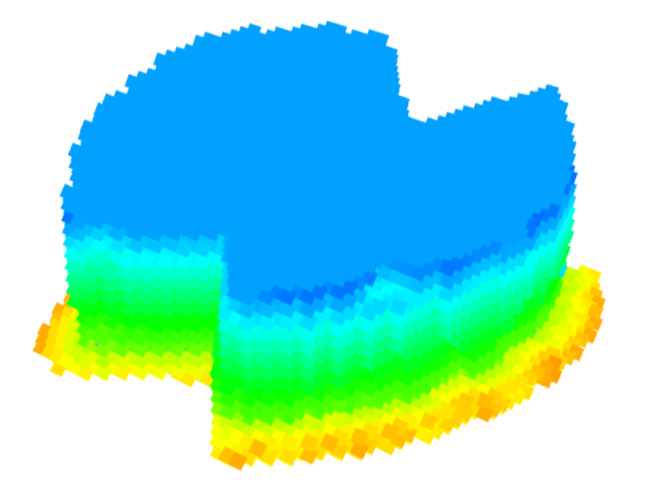}}
    \caption{Effect of pointcloud filtering in a two-UAV mapping scenario. (a) Experimental setup. (b,c) TSDF visualization from the left and right UAV perspectives using unfiltered pointclouds, where UAVs are reconstructed as obstacles. The highlighted orange regions indicate areas of mutual mapping. (d,e) TSDF visualization with pointcloud filtering enabled, where inter-robot reconstruction is removed, and only the static environment remains.}
    \label{fig:multi_filtering}
\end{figure}

Figure~\ref{fig:multi_filtering} illustrates the effect of this filtering process in a controlled two-UAV scenario. Two UAVs are positioned four meters apart and oriented toward each other in an environment containing only a ground plane, visualized as a purple mesh generated from the TSDF layer (Fig.~\ref{fig:drones_setup}). The TSDF layer of the centralized Voxblox map is visualized from the perspective of each UAV, both without and with pointcloud filtering.

Voxels are colored according to their stored signed distance values: distances beyond the maximum truncation distance are shown in blue, while smaller distances are progressively represented as cyan, green, yellow, and orange, with orange corresponding to voxels closest to a surface. Under this color scheme, the ground plane appears predominantly in orange. Without filtering (Figs.~\ref{fig:unfiltered_left} and~\ref{fig:unfiltered_right}), additional cyan and green regions appear in a spherical volume around each UAV, corresponding to observations of the other UAV and indicating mutual reconstruction within the map. When filtering is applied (Figs.~\ref{fig:filtered_left} and~\ref{fig:filtered_right}), these regions are no longer present, and the TSDF reflects only the distance to the static ground plane. This demonstrates that pointcloud filtering prevents inter-robot reconstruction and preserves a correct representation of free space for motion planning.

\section{EXPERIMENTAL EVALUATION} \label{sec:experimental_evaluation}

The proposed centralized framework is evaluated in the simulation environment of \cite{Mendes_2026}, using the MRS UAV system \cite{Baca_2021}, extended here to support a team of three DJI F450 UAVs. Each UAV is equipped with a simulated Intel RealSense D435i depth camera and shares a centralized Voxblox map. State estimation and trajectory tracking are handled by the MRS UAV system using a simulated RTK-GPS and Model Predictive Controller (MPC), respectively. 

In line with standard practice in exploration literature \cite{Bircher_2016, Selin_2019, Mendes_2026}, we decouple the exploration problem from localization by assuming sufficiently accurate state estimation, provided here by the simulated RTK-GPS. This allows us to isolate the impact of the proposed multi-UAV system on exploration performance without introducing extra errors from state estimation. In practical deployments, this assumption can be satisfied using high-precision localization systems such as RTK-GPS in outdoor environments or robust multi-agent Simultaneous Localization and Mapping (SLAM) pipelines in GPS-denied scenarios. The proposed framework can operate with any localization backend, as long as pose estimates are sufficiently consistent to support TSDF integration.

We evaluate the performance of four sampling-based planners - RH-NBVP \cite{Bircher_2016}, AEP \cite{Selin_2019}, KRH-NBVP \cite{Mendes_2026}, and KAEP \cite{Mendes_2026} - integrated into our multi-robot framework via system architecture adaptations. The experiments are conducted across three environments with different spatial features: a Maze ($20 \times 18 \times 2.5\,\mathrm{m}$) designed to test navigation in narrow passages; a Police Station ($20 \times 20 \times 15\,\mathrm{m}$) representing a compact reconstruction in an open-air environment; and a School ($50 \times 35 \times 20\,\mathrm{m}$) representing a large-scale reconstruction task with more complex geometry.

\begin{figure} [t]
  \centering
  \subfloat[Starting positions in maze.]{
    \label{subfig:start_maze_environment}
    \includegraphics[width=0.28\columnwidth]{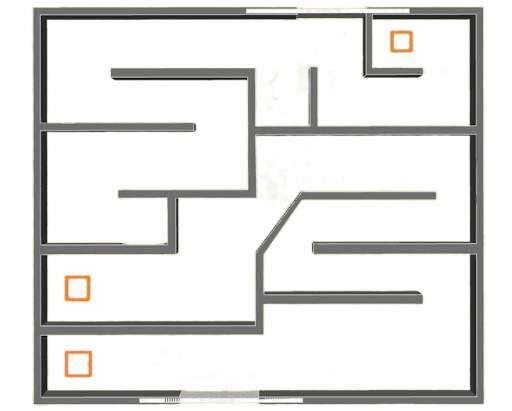}}
  \subfloat[Starting positions in police station.]{
    \label{subfig:start_police_environment}
    \includegraphics[width=0.31\columnwidth]{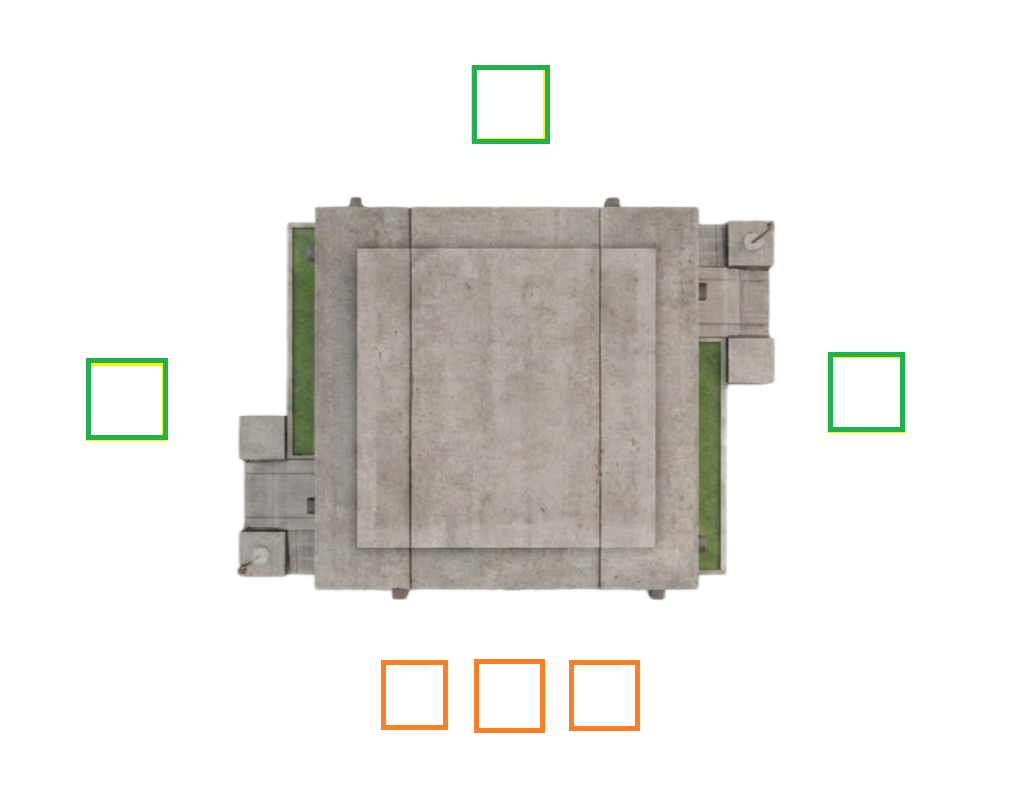}}
  \subfloat[Starting positions in school.]{
    \label{subfig:start_school_environment}
    \includegraphics[width=0.31\columnwidth]{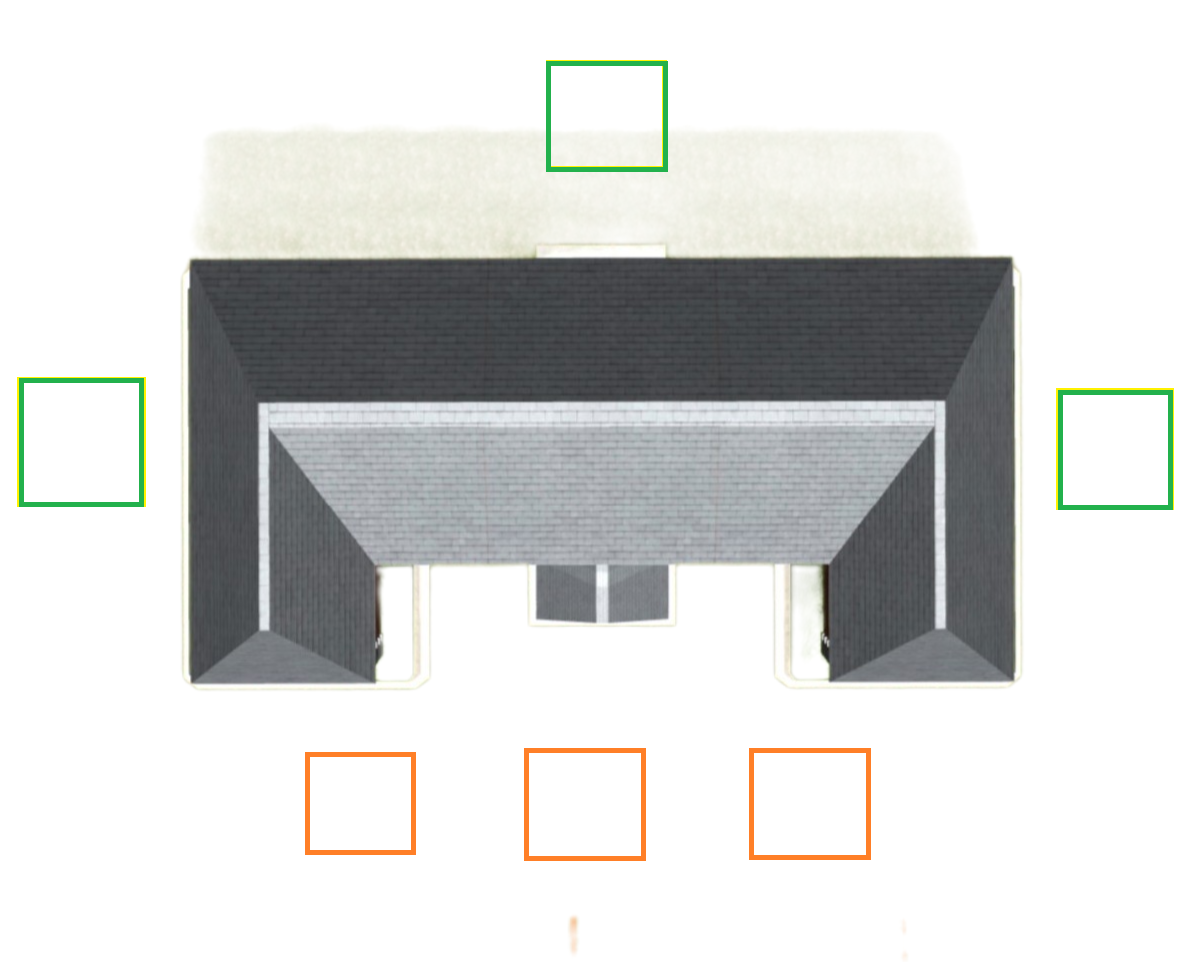}}
  \caption{UAV initialization layouts in maze (left), police station (middle), and school (right) environments for a team of three UAVs. The green boxes represent the SS configurations, and the orange boxes represent the JS configurations.}\label{fig:start_environments}
\end{figure}

Two initialization strategies are analyzed (Fig.~\ref{fig:start_environments}): a joint start (JS), where all UAVs are initialized in the same region, and a separated start (SS), where UAVs start in different regions of the environment. In the maze scenario, only SS is evaluated due to the narrow passages and the presence of a single feasible traversal path. A joint start would force the UAVs into single-file motion, resulting in redundant and inefficient exploration. Figure~\ref{subfig:start_maze_environment} illustrates the SS configuration in the maze for a team of three UAVs, where they are initialized near the beginning, middle, and end of the maze. For the police station and school environments, both JS and SS configurations are evaluated (Figs.~\ref{subfig:start_police_environment} and~\ref{subfig:start_school_environment}). In the SS configuration, UAVs are initialized in distinct regions (green boxes) to promote early task distribution, while in the JS configuration, all UAVs start from the same region (orange boxes), leading to initially overlapping exploration.

Performance is evaluated using four metrics: exploration rate, indicating the percentage of the environment explored over time; Full Exploration (FE), measuring the completeness of the 3D reconstruction at the end of an experiment; total path length (PL), reflecting path efficiency; and average velocity (AV), indicating how efficiently the UAVs utilize their physical capabilities. Exploration rate is reported at 25\%, 50\%, and 95\% completion (EX\%). In a multi-UAV context, PL represents the sum of the traveled distances of all UAVs, while AV is computed as the mean of the individual UAV average velocities. Each experiment is repeated five times, and the mean and standard deviation are reported.

All simulation parameters, planner settings, and sensor configurations are identical to those used in the single-UAV study of \cite{Mendes_2026}. No parameters were modified or re-tuned for the proposed multi-UAV framework in order to preserve a direct and fair comparison. The only additions concern the deployment configuration (number of UAVs and initialization strategy), which is unique to this work. For completeness, the most relevant parameters are summarized below in Table~\ref{tab:common_parameters}. As in \cite{Mendes_2026}, tighter constraints are used in the maze environment due to narrow passages, with $d_{replan}=\SI{0.4}{\meter}$ and $d_{collision}=\SI{0.7}{\meter}$. The full set can be found in \cite{Mendes_2026}.
\begin{table}[t]
\vspace*{2pt}
\caption{Key Experimental Parameters (unchanged from \cite{Mendes_2026}).}
\label{tab:common_parameters}
\begin{center}
\begin{tabularx}{0.98\linewidth}{>{\raggedright\arraybackslash}X r >{\raggedright\arraybackslash}X r}
\toprule
\textbf{Parameter} & \textbf{Value} & \textbf{Parameter} & \textbf{Value} \\
\midrule
$v_{xy_{\max}}, v_{z_{\max}}$ & \SI{1}{\meter\per\second}
& FoV & [$87^\circ$, $58^\circ$] \\
$\dot{\psi}_{\max}$ & \SI{2}{\radian\per\second} 
& Camera range & \SI{5}{\meter} \\
$a_{xy_{\max}}, a_{z_{\max}}$ & \SI{1}{\meter\per\second\squared} 
& Camera pitch & \SI{10}{\degree} \\
$\ddot{\psi}_{\max}$ & \SI{2}{\radian\per\second\squared} 
& $\mathbf{v}_{size}$ & \SI{0.2}{\meter} \\
$d_{replan}$ & \SI{0.8}{\meter} 
& $d_{collision}$ & \SI{1.5}{\meter} \\
\bottomrule
\end{tabularx}
\end{center}
\end{table}

\subsection{Multi-Robot Map Fusion Accuracy and Performance}
\label{subsec:map_performance}

To validate the consistency and computational performance of the proposed centralized architecture, we evaluate both the accuracy of the centralized map and the processing times for local sub-map integration and centralized map merging. All experiments were executed on a workstation equipped with an Intel Core i9 (14th Gen) CPU, 32 GB of RAM, and an NVIDIA GeForce RTX 4060 GPU. Results for a three-UAV team, evaluated in the police station and school environment using the KAEP planner under the JS configuration, are summarized in Table \ref{tab:compute_metrics}.

Map accuracy is measured by comparing the final centralized TSDF layer against the ground-truth mesh of the environments. As shown in Table \ref{tab:compute_metrics}, the centralized map achieves a low Root Mean Square Error (RMSE) across both environments. The observed RMSE remains on the order of the voxel resolution ($\mathbf{v}_{size} = \SI{0.2}{\meter}$), indicating that the fusion process preserves the overall geometric structure and is suitable for motion planning.

From a computational perspective, the framework distributes the mapping workload between UAVs and the central server. High-frequency pointcloud integration is performed onboard each UAV, with local TSDF and ESDF updates averaging 87.0 ms and 10.0 ms per update in the school environment. Since the same workstation simultaneously handles the physics simulation, multiple UAV mapping pipelines, and the central server, these local integration times represent a conservative upper bound. In contrast, the centralized server processes only incremental sub-map updates, resulting in significantly lower fusion times (8.6 ms for TSDF and 10.1 ms for ESDF in the School environment).

These results demonstrate that the proposed framework manages to offload dense mapping computations to the UAVs, preventing the central server from becoming a computational bottleneck and enabling real-time multi-UAV exploration.

\begin{table}[t]
\vspace*{2pt}
\caption{Computational Performance and Map Accuracy for KAEP in three-UAV Team.}
\label{tab:compute_metrics}
\begin{center}
\small 
\setlength{\tabcolsep}{4pt} 
\begin{tabularx}{0.98\linewidth}{@{} l >{\centering\arraybackslash}X >{\centering\arraybackslash}X @{}}
\toprule
\textbf{Metric} & \textbf{Police Station} & \textbf{School} \\
\midrule
Local TSDF Integration (ms) & 113.0 $\pm$ 15.4 & 87.0 $\pm$ 13.1 \\
Local ESDF Integration (ms) & 9.2 $\pm$ 8.2 & 10.0 $\pm$ 4.7 \\
Central TSDF Merge (ms) & 7.9 $\pm$ 2.6 & 8.6 $\pm$ 8.5 \\
Central ESDF Merge (ms) & 8.9 $\pm$ 1.9 & 10.1 $\pm$ 9.2 \\
Central Map RMSE (m) & 0.24 $\pm$ 0.20 & 0.19 $\pm$ 0.15 \\
\bottomrule
\end{tabularx}
\end{center}
\end{table}

\subsection{Simulated Experiments}\label{subsec:simulated_experiments}

\begin{figure*} [t]
  \centering
  \subfloat[Exploration progress of Maze using the SS Approach.]{
    \label{subfig:multi_maze_exp_rate}
    \includegraphics[width=0.65\columnwidth]{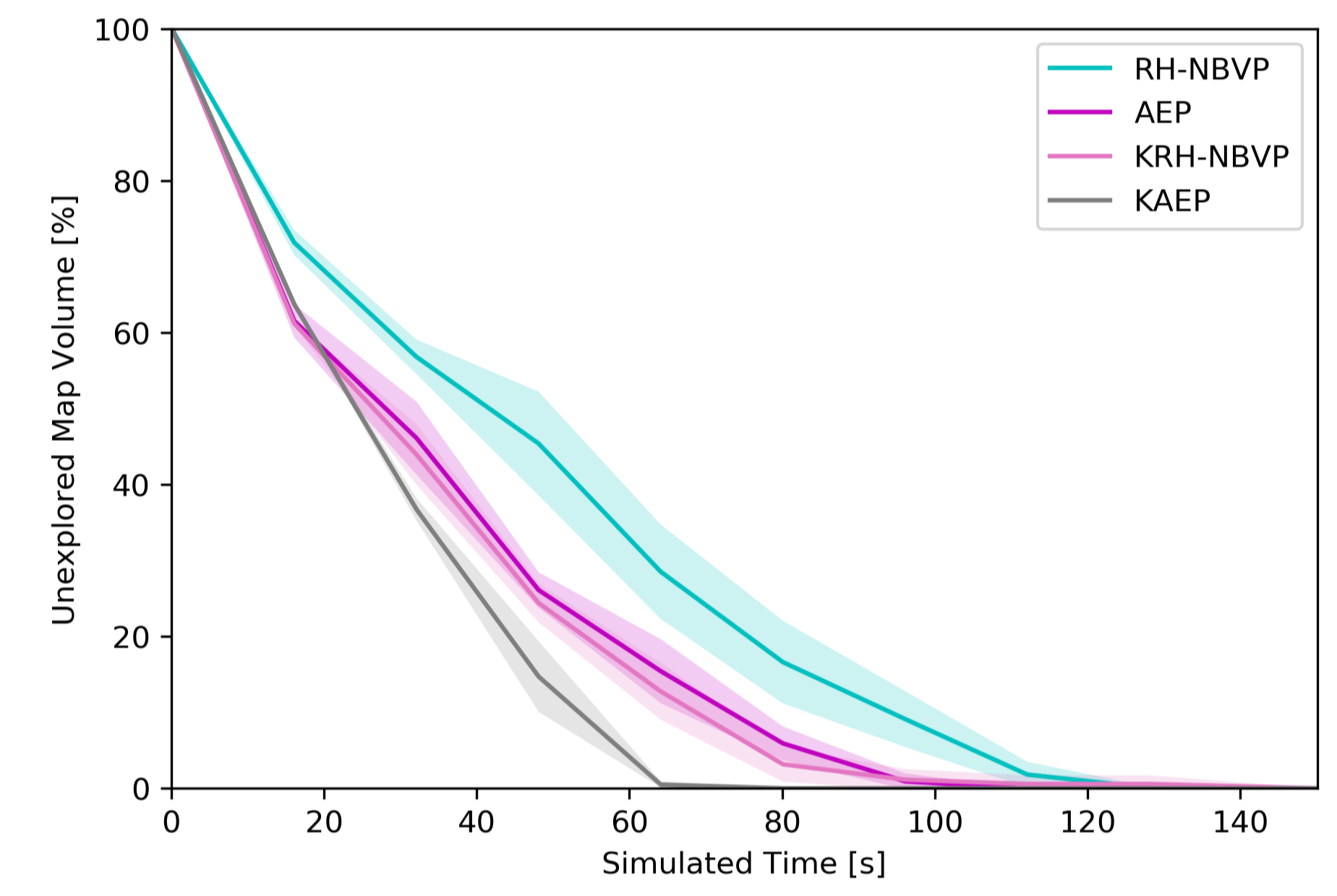}}
  \hspace{\stretch{0.1}}%
  \subfloat[Exploration progress of Police Station using the SS Approach.]{
    \label{subfig:multi_school_exp_rate_JS}
    \includegraphics[width=0.65\columnwidth]{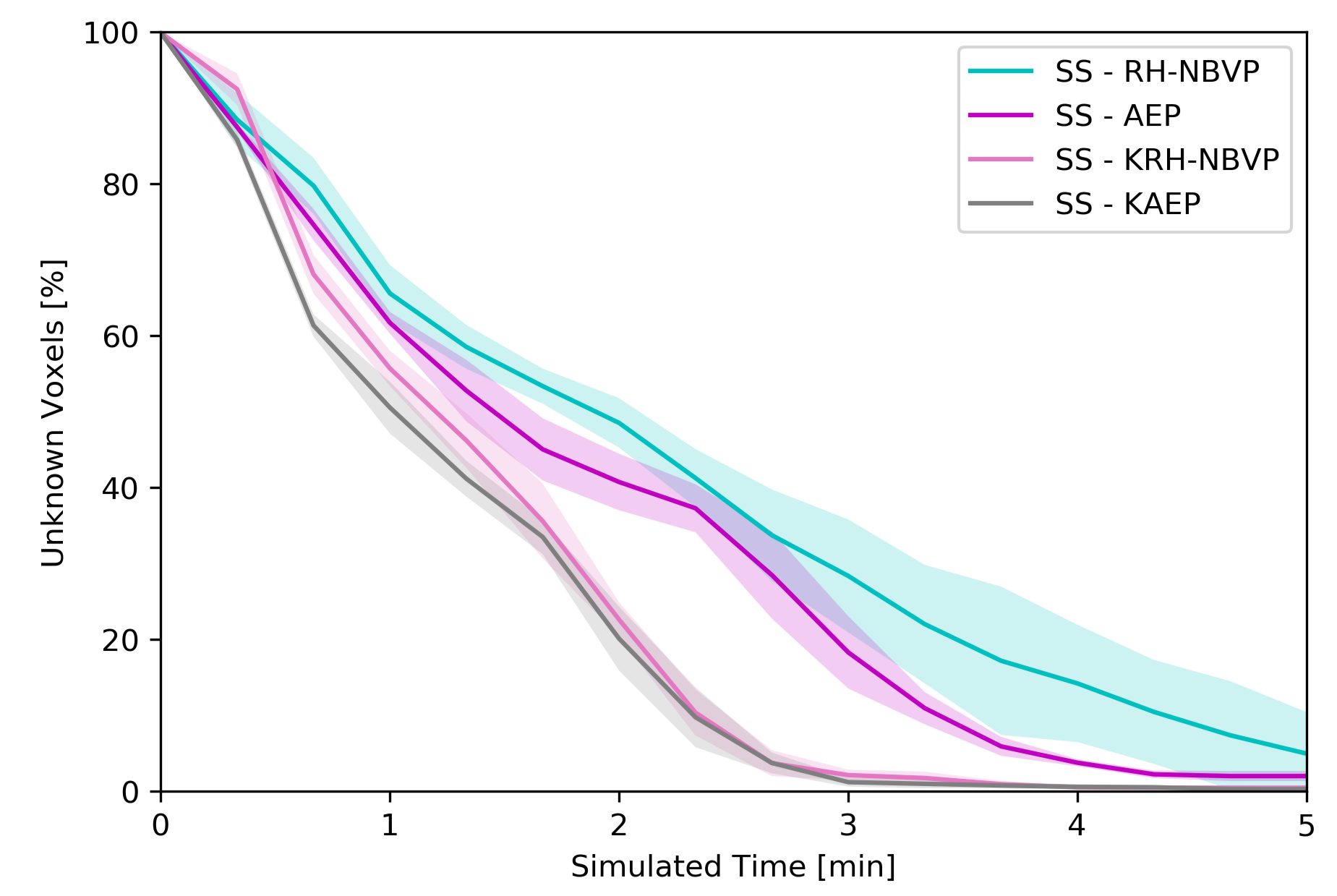}}
  \hspace{\stretch{0.1}}%
  \subfloat[Exploration progress of School using the SS Approach.]{
    \label{subfig:multi_school_exp_rate_SS}
    \includegraphics[width=0.65\columnwidth]{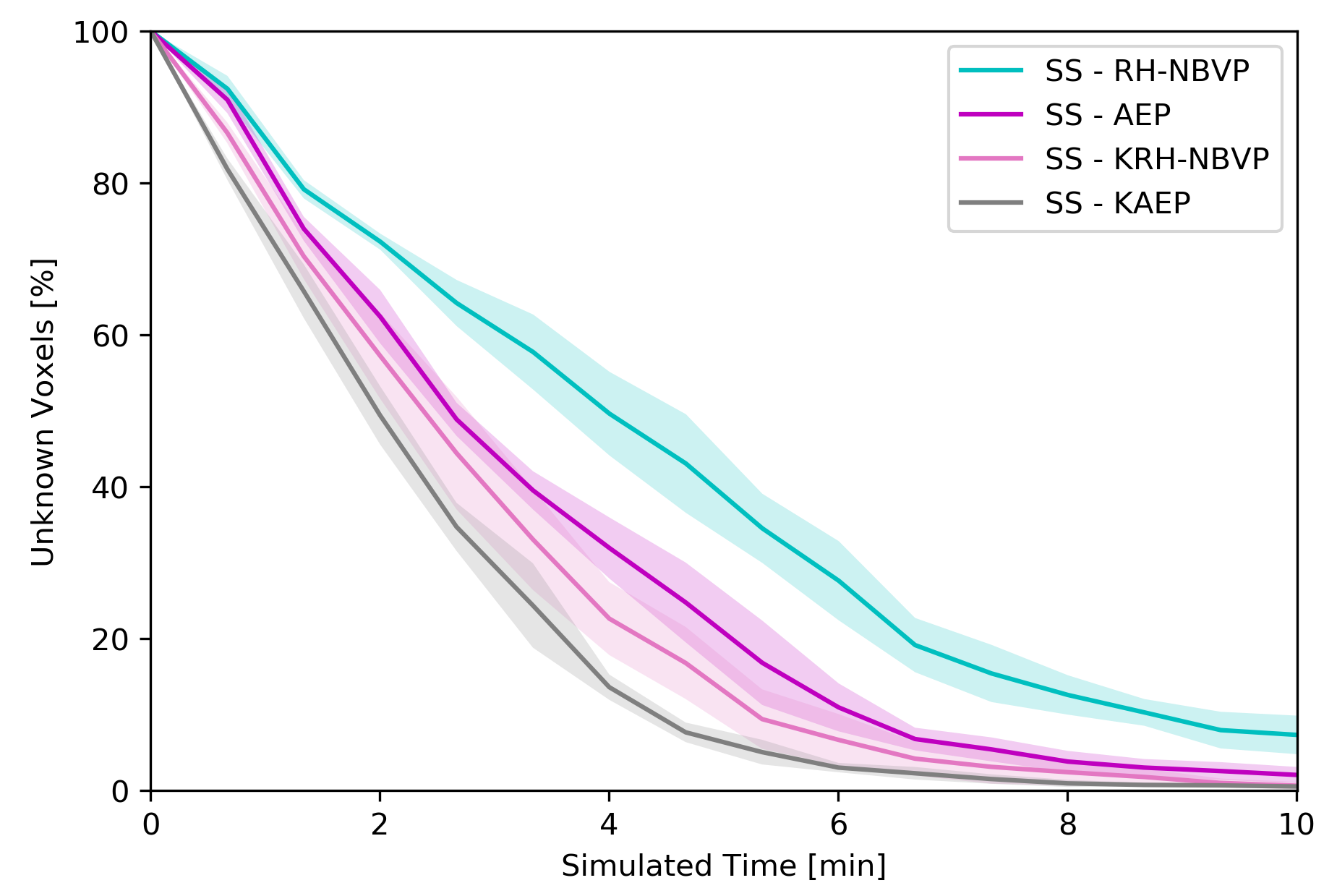}}
  \vspace{0.5cm}
  \subfloat[Exploration progress of Police Station using the JS Approach.]{
    \label{subfig:multi_police_exp_rate_JS}
    \includegraphics[width=0.65\columnwidth]{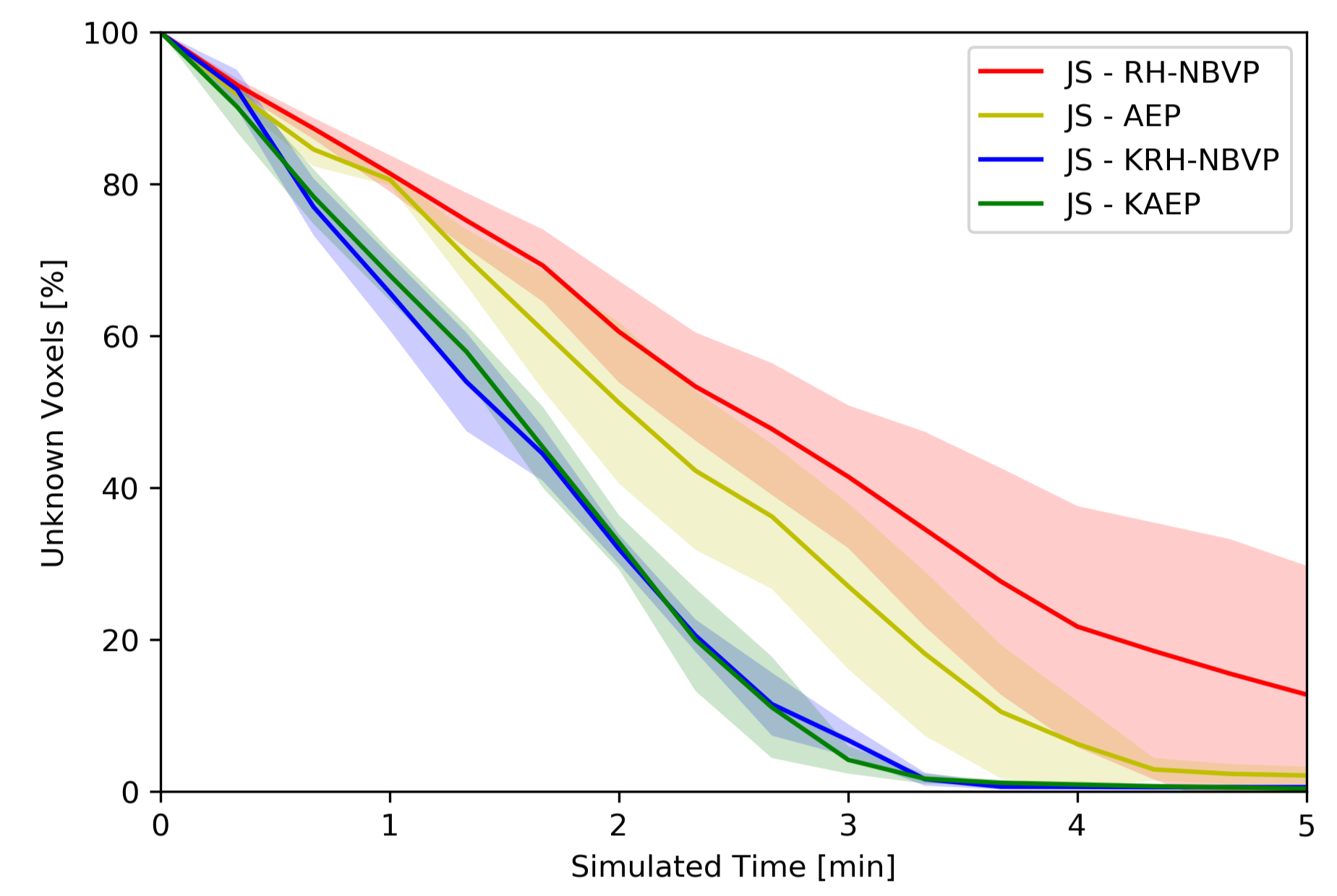}}
  \subfloat[Exploration progress of School using the JS Approach.]{
    \label{subfig:multi_police_exp_rate_SS}
    \includegraphics[width=0.65\columnwidth]{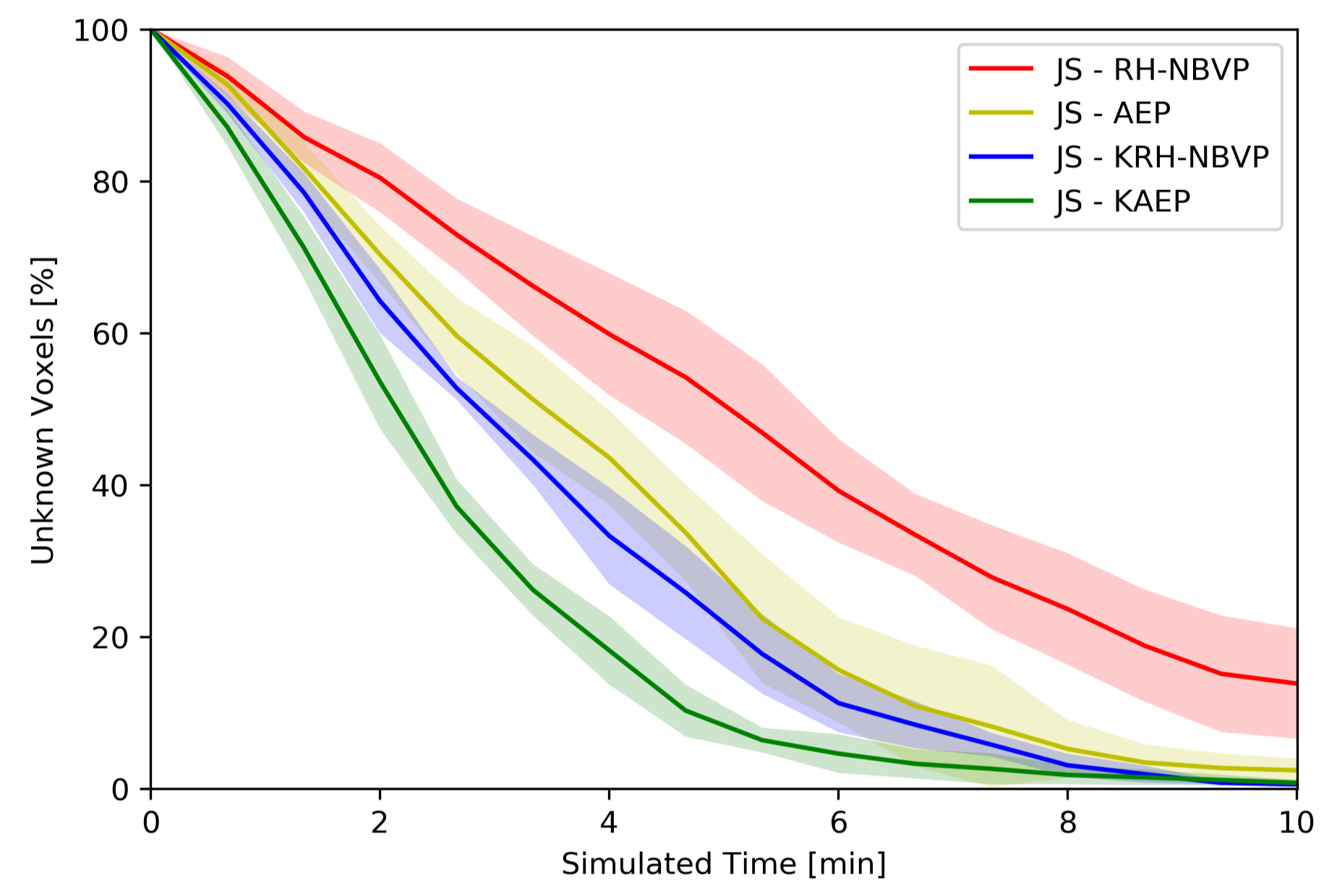}}
  \caption{Exploration progress results for maze, police station, and school environments using a three-UAV team under SS (top) and JS (bottom) deployment strategies.}\label{fig:multi_exp_rate}
\end{figure*}

\begin{table*}[t]
\vspace*{2pt}
\setlength{\tabcolsep}{3pt}
\renewcommand{\arraystretch}{1.1}
\caption{Combined Multi-Robot Experiment Results. Note: Time units vary by environment. The dashes (--) indicate that E95\% exploration was not reached before the simulation time limit.}
\label{tab:combined_results}
\centering
\scriptsize 
\begin{tabularx}{0.98\linewidth}{c l l 
    >{\centering\arraybackslash}X 
    >{\centering\arraybackslash}X 
    >{\centering\arraybackslash}X 
    >{\centering\arraybackslash}X 
    >{\centering\arraybackslash}X 
    >{\centering\arraybackslash}X}
\toprule
\textbf{Environment} & \textbf{Cfg} & \textbf{Planner} & \textbf{E25\%} & \textbf{E50\%} & \textbf{E95\%} & \textbf{FE} & \textbf{PL} & \textbf{AV} \\
& & & Time & Time & Time & [\SI{}{\percent}] & [\SI{}{\meter}] & [\SI{}{\meter\per\second}] \\
\midrule

\multirow{4}{*}{\shortstack[c]{\textbf{Maze}\\(Time in [\SI{}{\second}])}} 
& \multirow{4}{*}{SS} 
  & RH-NBVP  & $14.32 \pm 0.88$ & $41.67 \pm 8.48$ & $105.11 \pm 7.72$ & $\pmb{100.0 \pm 0.0}$ & $\pmb{236.75 \pm 3.52}$ & $0.27 \pm 0.01$ \\
& & AEP      & $10.51 \pm 0.62$ & $28.09 \pm 4.68$ & $83.02 \pm 5.71$  & $\pmb{100.0 \pm 0.0}$ & $275.58 \pm 29.90$ & $0.29 \pm 0.03$ \\
& & KRH-NBVP & $\pmb{10.42 \pm 0.08}$ & $26.51 \pm 3.19$ & $77.04 \pm 5.54$  & $\pmb{100.0 \pm 0.0}$ & $306.48 \pm 1.99$ & $0.35 \pm 0.01$ \\
& & KAEP     & $11.15 \pm 0.04$ & $\pmb{24.26 \pm 0.50}$ & $\pmb{59.04 \pm 2.81}$ & $\pmb{100.0 \pm 0.0}$ & $348.92 \pm 29.46$ & $\pmb{0.42 \pm 0.02}$ \\
\midrule

\multirow{8}{*}{\shortstack[c]{\textbf{Police}\\\textbf{Station}\\(Time in [\SI{}{\minute}])}} 
& \multirow{4}{*}{JS}
  & RH-NBVP  & $1.35 \pm 0.25$ & $2.53 \pm 0.55$ & -- & $87.24 \pm 16.97$ & $\pmb{433.93 \pm 92.69}$ & $0.45 \pm 0.10$ \\
& & AEP      & $1.18 \pm 0.11$ & $2.05 \pm 0.42$ & $4.13 \pm 0.65$ & $97.89 \pm 1.17$ & $576.25 \pm 42.35$ & $0.60 \pm 0.04$ \\
& & KRH-NBVP & $\pmb{0.73 \pm 0.13}$ & $\pmb{1.47 \pm 0.20}$ & $3.12 \pm 0.16$ & $99.43 \pm 0.37$ & $531.43 \pm 30.22$ & $0.75 \pm 0.01$ \\
& & KAEP     & $0.78 \pm 0.11$ & $1.55 \pm 0.14$ & $\pmb{2.96 \pm 0.32}$ & $\pmb{99.62 \pm 0.24}$ & $828.21 \pm 35.50$ & $\pmb{0.85 \pm 0.04}$ \\
\cmidrule{2-9} 
& \multirow{4}{*}{SS}
  & RH-NBVP  & $0.78 \pm 0.09$ & $1.90 \pm 0.19$ & -- & $94.97 \pm 5.47$ & $\pmb{482.15 \pm 8.73}$ & $0.49 \pm 0.01$ \\
& & AEP      & $0.66 \pm 0.05$ & $1.45 \pm 0.18$ & $3.81 \pm 0.17$ & $97.99 \pm 0.66$ & $546.32 \pm 29.54$ & $0.60 \pm 0.03$ \\
& & KRH-NBVP & $0.57 \pm 0.03$ & $1.20 \pm 0.13$ & $\pmb{2.60 \pm 0.12}$ & $99.53 \pm 0.15$ & $546.39 \pm 38.92$ & $0.76 \pm 0.02$ \\
& & KAEP     & $\pmb{0.48 \pm 0.02}$ & $\pmb{1.02 \pm 0.11}$ & $\pmb{2.60 \pm 0.18}$ & $\pmb{99.72 \pm 0.12}$ & $855.29 \pm 14.18$ & $\pmb{0.88 \pm 0.01}$ \\
\midrule

\multirow{8}{*}{\shortstack[c]{\textbf{School}\\(Time in [\SI{}{\minute}])}} 
& \multirow{4}{*}{JS}
  & RH-NBVP  & $2.49 \pm 0.54$ & $5.05 \pm 0.86$ & -- & $86.16 \pm 7.27$ & $\pmb{960.05 \pm 17.44}$ & $0.52 \pm 0.01$ \\
& & AEP      & $1.73 \pm 0.22$ & $3.45 \pm 0.54$ & $8.09 \pm 1.66$ & $97.58 \pm 1.54$ & $1276.67 \pm 34.13$ & $0.70 \pm 0.02$ \\
& & KRH-NBVP & $1.50 \pm 0.16$ & $2.86 \pm 0.18$ & $7.53 \pm 0.65$ & $99.44 \pm 0.25$ & $1328.81 \pm 14.72$ & $0.79 \pm 0.01$ \\
& & KAEP     & $\pmb{1.18 \pm 0.17}$ & $\pmb{2.15 \pm 0.23}$ & $\pmb{5.86 \pm 1.06}$ & $\pmb{99.45 \pm 0.41}$ & $1512.20 \pm 265.57$ & $\pmb{0.84 \pm 0.15}$ \\
\cmidrule{2-9}
& \multirow{4}{*}{SS}
  & RH-NBVP  & $1.74 \pm 0.11$ & $3.97 \pm 0.64$ & -- & $92.67 \pm 2.53$ & $\pmb{950.18 \pm 72.76}$ & $0.52 \pm 0.04$ \\
& & AEP      & $1.30 \pm 0.08$ & $2.61 \pm 0.13$ & $7.51 \pm 0.71$ & $97.95 \pm 1.05$ & $1267.90 \pm 8.06$ & $0.70 \pm 0.01$ \\
& & KRH-NBVP & $1.15 \pm 0.12$ & $2.38 \pm 0.39$ & $6.45 \pm 1.08$ & $99.36 \pm 0.48$ & $1310.67 \pm 46.57$ & $0.80 \pm 0.03$ \\
& & KAEP     & $\pmb{0.95 \pm 0.11}$ & $\pmb{1.98 \pm 0.17}$ & $\pmb{5.35 \pm 0.37}$ & $\pmb{99.47 \pm 0.15}$ & $1581.34 \pm 98.22$ & $\pmb{0.87 \pm 0.05}$ \\
\bottomrule
\end{tabularx}
\end{table*}

The quantitative results are summarized in Fig.~\ref{fig:multi_exp_rate} and Table~\ref{tab:combined_results}. Across all environments and deployment strategies, KAEP consistently achieves the fastest exploration rates, followed by KRH-NBVP, AEP, and RH-NBVP. This trend is observed throughout the exploration process, including early exploration stages (E25\% and E50\%) and near-complete coverage (E95\%).

As illustrated by the reconstructions in Figure \ref{fig:map_reconstruction_comparison}, the deployment strategy significantly impacts exploration speed. In all environments, the SS configuration results in faster exploration compared to the JS configuration across all exploration stages (E25\%, E50\%, and E95\%). This effect becomes more pronounced as the environment size increases, where SS deployments reduce early-stage overlap by allowing UAVs to initially cover spatially distinct regions. In contrast, JS deployments lead to redundant reconstruction during the initial exploration phase, delaying overall coverage.

Relative to trends reported in prior single-UAV exploration experiments~\cite{Mendes_2026}, the multi-UAV setup improves the robustness of sampling-based planners, particularly RH-NBVP and KRH-NBVP. This is because these two planners are prone to getting stuck in local minima due to their lack of a global planning mechanism. In the maze environment, single-UAV planners that previously got stuck in local minima are able to achieve full exploration when multiple UAVs are deployed. This improvement arises from the implicit spatial task allocation using the SS configuration, which distributes the exploration effort without requiring explicit inter-robot coordination.

The average velocity trends observed in the multi-robot experiments are consistent across all environments and planners. KAEP achieves the highest average velocity, followed by KRH-NBVP, AEP, and RH-NBVP. However, average velocities in the multi-UAV system are generally lower than those observed in single-UAV experiments~\cite{Mendes_2026}, particularly in the maze environment. This is a result of task distribution within the environment, often requiring the UAVs to move in a back-and-forth motion within their exploration section.

Overall, these results demonstrate that centralized multi-UAV exploration not only accelerates coverage but also mitigates failure modes observed in single-UAV exploration, with the choice of initial deployment playing a significant role in overall system performance.

\section{CONCLUSION} \label{sec:conclusion}

In this paper, we presented a centralized multi-UAV exploration framework that extends established single-UAV sampling-based exploration planners to a multi-UAV setting. The proposed architecture addresses three fundamental challenges: fusing data from multiple sensors into a consistent global map, ensuring safety against environmental and inter-UAV collisions, and preventing the reconstruction of other UAVs within the static map.

To achieve this, the framework builds upon the voxblox mapping library and adapts it to support centralized multi-robot operation, allowing multiple UAVs to contribute to a shared TSDF map. Furthermore, an inter-UAV collision avoidance module was adapted, and a pointcloud filtering module was developed to operate within this centralized planning pipeline. These components ensure that UAVs can navigate safely and build accurate maps without introducing dynamic obstacles that would otherwise restrict the planning process.

The framework was evaluated using four sampling-based planners - RH-NBVP, KRH-NBVP, AEP, and KAEP - under different deployment configurations. Results show that initializing UAVs from SS configurations leads to faster exploration and improved coverage compared to JS deployments. These findings highlight that initial deployment strategies play a critical role in multi-UAV exploration performance, even when using identical planning algorithms and without introducing explicit task allocation or inter-robot coordination mechanisms.

While the centralized approach facilitates consistent map fusion and conflict-aware coordination, it introduces a single point of failure and limits scalability as the number of UAVs increases. These trade-offs motivate future work in two directions: first, validate the proposed centralized framework in real-world multi-UAV experiments, and second, extend the architecture toward a decentralized planning system to improve the robustness and scalability in larger fleets.

\bibliographystyle{IEEEtran}
\bibliography{references}

\end{document}